\documentclass[letterpaper, 10 pt, conference]{ieeeconf}  % Comment this line out if you need a4paper

\IEEEoverridecommandlockouts                              % This command is only needed if 
\usepackage{graphicx}
\usepackage{graphics} % for pdf, bitmapped graphics files
\usepackage{epsfig} % for postscript graphics files
\usepackage{times} % assumes new font selection scheme installed
\usepackage[hidelinks]{hyperref}
\usepackage{authblk}
\usepackage{amsmath} % assumes amsmath package installed
\usepackage{amssymb}  % assumes amsmath package installed
\usepackage{booktabs}
\usepackage{float}
\usepackage{siunitx}

\usepackage{todonotes}

\DeclareMathAlphabet{\mathcal}{OMS}{cmsy}{m}{n}

\title{RAPTOR: Rapid Aerial Pickup and Transport of Objects by Robots}

\author{
Aurel X. Appius$^{\dagger*}$, Erik Bauer$^{\dagger*}$, Marc Blöchlinger$^{\dagger*}$, Aashi Kalra$^{\dagger*}$, Robin Oberson$^{\dagger*}$, Arman Raayatsanati$^{\dagger*}$,\\ Pascal Strauch$^{\dagger*}$, Sarath Suresh$^{\dagger*}$, Marco von Salis$^{\dagger*}$, and
Robert K. Katzschmann$^{\dagger}$%
\thanks{$^{*}$Equal Contribution}
\thanks{$^{\dagger}$Soft Robotics Lab, ETH Zurich, Switzerland}
\thanks{\tt \footnotesize 
\{\href{mailto:appiusa@ethz.ch}{appiusa},
\href{mailto:erbauer@ethz.ch}{erbauer},
\href{mailto:mbloechli@ethz.ch}{mbloechli},
\href{mailto:akalra@ethz.ch}{akalra},
\href{mailto:roberson@ethz.ch}{roberson},
\href{mailto:araayatsa@ethz.ch}{araayatsa},
\href{mailto:pstrauch@ethz.ch}{pstrauch},
\href{mailto:ssuresh@ethz.ch}{ssuresh},
\href{mailto:mvonsalis@ethz.ch}{mvonsalis},
\href{mailto:rkk@ethz.ch}{rkk}\}@ethz.ch}
}

\begin{document}

\maketitle
%%%%%%%%%%%%%%%%%%%%%%%%%%%%%%%%%%%%%%%%%%%%%%%%%%%%%%%%%%%%%%%%%%%%%%%%%%%%%%%%

\begin{abstract}
Rapid aerial grasping through robots can lead to many applications that utilize fast and dynamic picking and placing of objects. Rigid grippers traditionally used in aerial manipulators require high precision and specific object geometries for successful grasping. We propose RAPTOR, a quadcopter platform combined with a custom Fin Ray\textsuperscript{\textregistered} gripper to enable more flexible grasping of objects with different geometries, leveraging the properties of soft materials to increase the contact surface between the gripper and the objects. To reduce the communication latency, we present a new lightweight middleware solution based on Fast DDS (Data Distribution Service) as an alternative to ROS (Robot Operating System). We show that RAPTOR achieves an average of 83\% grasping efficacy in a real-world setting for four different object geometries while moving at an average velocity of 1 m/s during grasping. In a high-velocity setting, RAPTOR supports up to four times the payload compared to previous works. Our results highlight the potential of aerial drones in automated warehouses and other manipulation applications where speed, swiftness, and robustness are essential while operating in hard-to-reach places.\footnote{Code: \url{https://github.com/raptor-ethz/raptor_setup}}
\end{abstract}

\section{Introduction}
Dynamic aerial grasping~\cite{Fishman2021DynamicGW} is a recent research challenge that bears much potential to enable many new applications in automation and hard to reach places. Advances in soft robotic grippers, namely the increased tolerance for tracking errors and additional damping when picking up objects, allow for grasping at higher speeds with mobile platforms such as multicopters. Multicopters and soft robotic grippers are a natural match for aerial manipulation as they are easy to maneuver and provide much versatility as a research platform. These properties have been beneficial for quadcopters in a wide range of applications such as point-to-point deliveries~\cite{Haque2014AutonomousQF} and aerial manipulation~\cite{Kamel2016DesignAM}. While there is extensive previous work on normal flights, recent approaches continue to incorporate new features of quadcopter platforms~\cite{Fishman2021DynamicGW,Bodie2019AnOA, Falanga2019TheFD}. 

We continue to push the boundaries of dynamic aerial grasping with RAPTOR---a flying robot consisting of a soft robotic gripper attached to a quadcopter. Our approach utilizes the maneuverability of the quadcopter platform while the soft gripper ensures robustness and adaptability to the object geometry and size. The resulting system is capable of grasping a variety of objects up to a total maximum payload of \SI{400}{\gram}. In addition, the quadcopter reaches average grasping velocities of up to \SI{1}{\m\per\s} and velocities of up to \SI{6}{\m\per\s} when not carrying a payload. 

\begin{figure}[t]
  \centering
  \includegraphics[width=\linewidth]{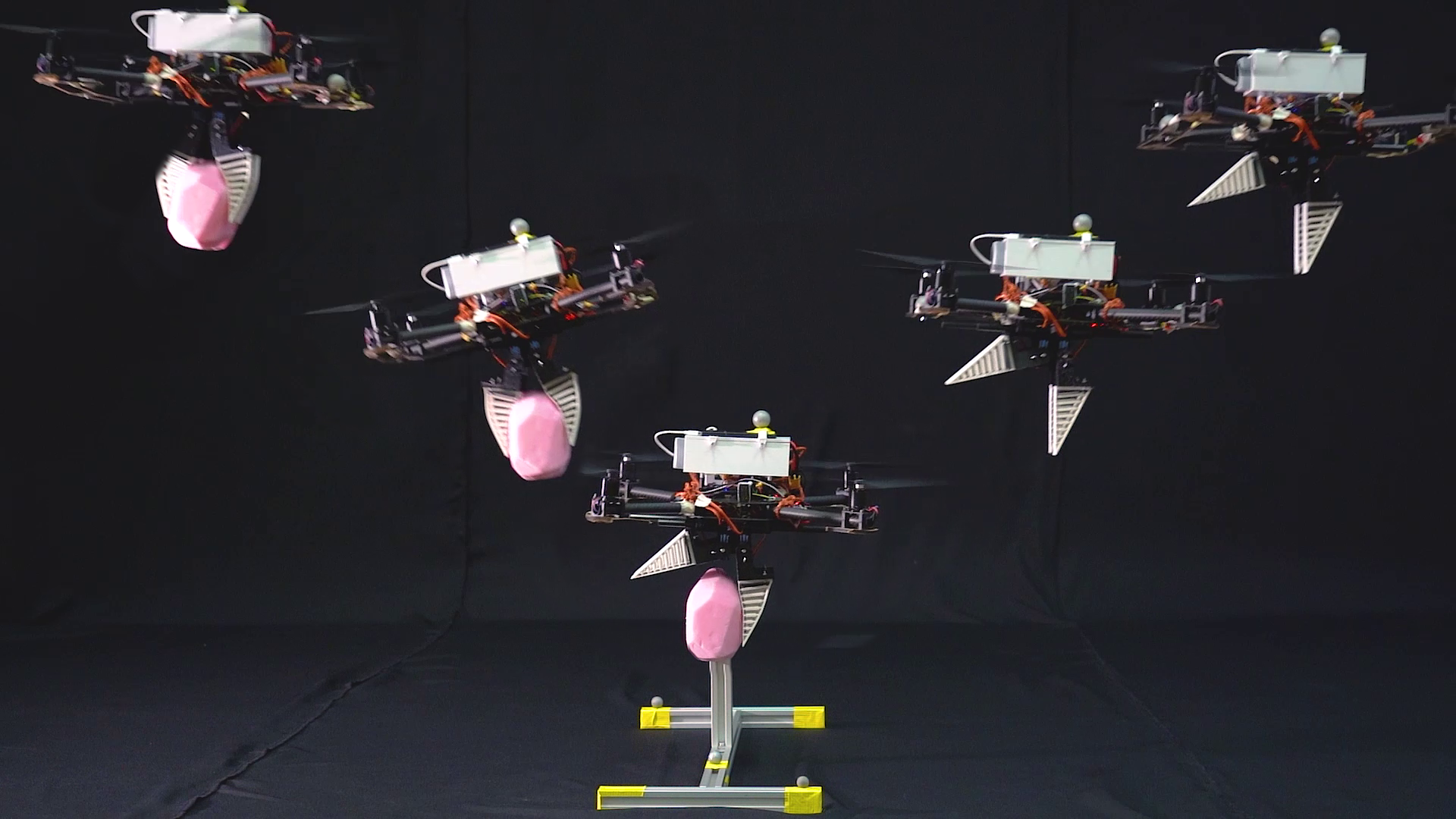}
  \centering
  \\
  \caption{The RAPTOR platform dynamically picks up an object in-flight. The compliant soft gripper is able to adapt to the irregular shape. A rigid gripper with the same design fails to pick up this object due to the gripper's low adaptability and compliance.}
\end{figure}

\subsection{Related Work}
Combining robotic grippers with flying platforms opens a new dimension for aerial object transfer. Thomas \textit{et al.}~\cite{Thomas2013AvianInspiredGF} have demonstrated the feasibility of this approach by picking up cylindrical objects with a quadcopter using an attached swing arm. Recent work shows the potential of using nonlinear model predictive control for a quadcopter in the context of aerial manipulation~\cite{Garimella2018NonlinearMP}. Moreover, related work explores high speed deliveries using a hook for high speed delivery~\cite{Tanaka2019HighspeedUD} and high speed grasping with passive rigid claws~\cite{Stewart2022HowTS}. The challenge of increasing robustness towards unknown object geometries and surfaces can be addressed with soft robotic grippers. 
Soft robotics~\cite{Rus2015DesignFA, GeorgeThuruthel2018ControlSF} has been a growing field of research, with many biologically inspired soft tendons and grippers being developed~\cite{King2018DesignFA,MantiMariangela2015ABS,Hassan2015DesignAD}. In particular, the discovery of the Fin Ray\textsuperscript{\textregistered} effect~\cite{Shan2020ModelingAA} opens up many possibilities for the application of soft grippers. In addition, related work shows that tilting the ribs~\cite{Ali2019BiologicallyIG} and combining hard and soft materials~\cite{Crooks2017PassiveGI} enhances gripping strength and proposes tendon-driven robot hands for high speed grasping~\cite{McLaren2019APC}.
Fishman \textit{et al.}~\cite{Fishman2021DynamicGW, Fishman2021ControlAT} propose a tendon-driven aerial gripper mounted on a quadcopter. This approach uses adaptive control methods to compensate for the unmodeled aerodynamics such as a ground effect that induces tracking errors. In their experiments, they report a 91.7\% success rate over a sample size of 23 grasping attempts with a grasping speed of \SI{0.2}{\m\per\s}. Our goal is to address some of the open challenges, particularly with regard to the robustness to object geometry and grasping at high velocities.

\subsection{Contribution}
In this paper, we propose a platform for aerial grasping with a significantly increased performance compared to previous works in rapid aerial grasping. We present an exhaustive description of the system architecture, including a novel approach to middleware using a custom Fast DDS (Data Distribution Service) wrapper as well as the electromechanical design of the drone platform and the gripper. Our Fast DDS wrapper is significantly faster compared to ROS (Robot Operating System) while being lightweight and cross-platform compatible.

In addition, we demonstrate the effectiveness of our system through real-world experiments that showcase high grasping speeds and robustness to different object geometries. Our platform shows significant improvements over previous works for rapid aerial grasping with RAPTOR and outlines the future research challenges for aerial manipulation.

\subsection{Outline}
Section~\ref{architecture} provides an overview of the system architecture with particular emphasis on the integration of the different subsystems. This is followed by a motivation of the mechanical design decisions in Section~\ref{mechanical}, describing the custom components we designed and manufactured as well as the modifications we made to off-the-shelf components. In Section~\ref{control}, we explore our trajectory generation and control strategies for robust trajectory tracking. Finally, we illustrate our results in Section~\ref{experiments} and show an exhaustive experimental validation of our platform and conclude the paper in Section~\ref{conclusion} with an outlook into future work.

\section{System Architecture}\label{architecture}
\subsection{System Components}
For subsystems like the drone, there already exist off-the-shelf components that allow us to compose our systems in a way to only develop custom components where needed. This allows us to maximize our efforts towards developing a robust system for grasping. We divide our system into three logical subsystems: 1) the drone, 2) the gripper module mounted on the drone, and 3) offboard components. 

\subsubsection{Drone Components}\label{Components}
\begin{itemize}
    \item Raspberry Pi 4B (Companion computer)
    \item Pixhawk 4 (Flight controller)
    \item XBee Pro Zigbee Module (RF communication directly to PX4 flight controller)
\end{itemize}

\subsubsection{Gripper Components}
\begin{itemize}
    \item Arduino Nano (Microcontroller for gripper control)
    \item Hobbywing BEC 3A (Step down LiPo voltage from 14.8V to 6V for servos)
    \item Two Megmoki MG995 Hi-Speed Servos (Gripper actuation)
\end{itemize}

\subsubsection{Offboard Components}
\begin{itemize}
    \item 19 Vicon MX-40 cameras (Motion capture data)
    \item Laptops/PCs (Trajectory generation and logging) 
\end{itemize}

\subsection{Middleware Design}
Robotic systems often use ROS~\cite{Quigley2009ROSAO} as middleware, as it provides an easy to use API and several bundled tools that are commonly used in robotics including for simulation, visualization, and mapping. However, the ROS1 transport system does not satisfy the temporal deadlines and fault tolerances required for real-time systems. This requirement is especially relevant for safety-critical systems like flying robots and autonomous vehicles. In addition, ROS1 is not a suitable choice for embedded systems due to its resource-intensive nature.

ROS2 addresses these issues by replacing the transport system with Fast DDS, an open-source industry standard for real-time communication. Fast DDS has proven successful in safety-critical applications such as aerospace, smart grids, defense, financial trading, and power plants. Underneath, it is a fully distributed system of high-performance point-to-point connections, avoiding data bottlenecks and single points of failure. The system continues to run even if individual nodes crash or restart. A key benefit of Fast DDS is that the discovery server handles all detection and configuration automatically when a new node is connected. These features allow the user to focus on transmitting structured data rather than the mechanics of communication.

ROS2 offers an API similar to ROS1, with support for multiple programming languages including C++ and Python. This means that messages must undergo format conversion twice: from ROS2 to Fast DDS while sending and from Fast DDS to ROS2 while receiving.  Consequently, two internal notifications are required per publishing/subscribing event. This increases the latency by a factor of up to 1.5~\cite{Kronauer2021LatencyOO}. Clearly, ROS2 is not the best choice for time-critical communication channels. On the other hand, native Fast DDS implementations provide the best performance, but they are much more complicated to use. We use a custom, lightweight wrapper over the native Fast DDS implementation that offers an API similar to ROS. To retain the same high-performance characteristics, we use only C++ abstractions and avoid message format conversions. The downside is that other programming languages are not directly supported. Communicating with Python programs is supported via bindings (pybind) at the cost of much higher latency. However, it is not an issue in our case since all time-critical functions are implemented in C++. Moreover, ROS2 is not officially supported on Raspberry Pi OS---the operating system for our onboard computer (Raspberry Pi). Our wrapper retains the cross-platform capability of native DDS implementations: Windows, macOS and most Linux distributions including Raspberry Pi OS are supported out of the box.

\subsubsection{Motion Capture System}
An overhead motion capture system tracks the pose of objects in the 3D space and publishes this information to the local network via Fast DDS. Although the system can run at frequencies higher than \SI{250}{\hertz}, we have found a frequency of \SI{100}{\hertz} to be sufficient.

\begin{figure}[h]
  \centering
  \includegraphics[width=\linewidth]{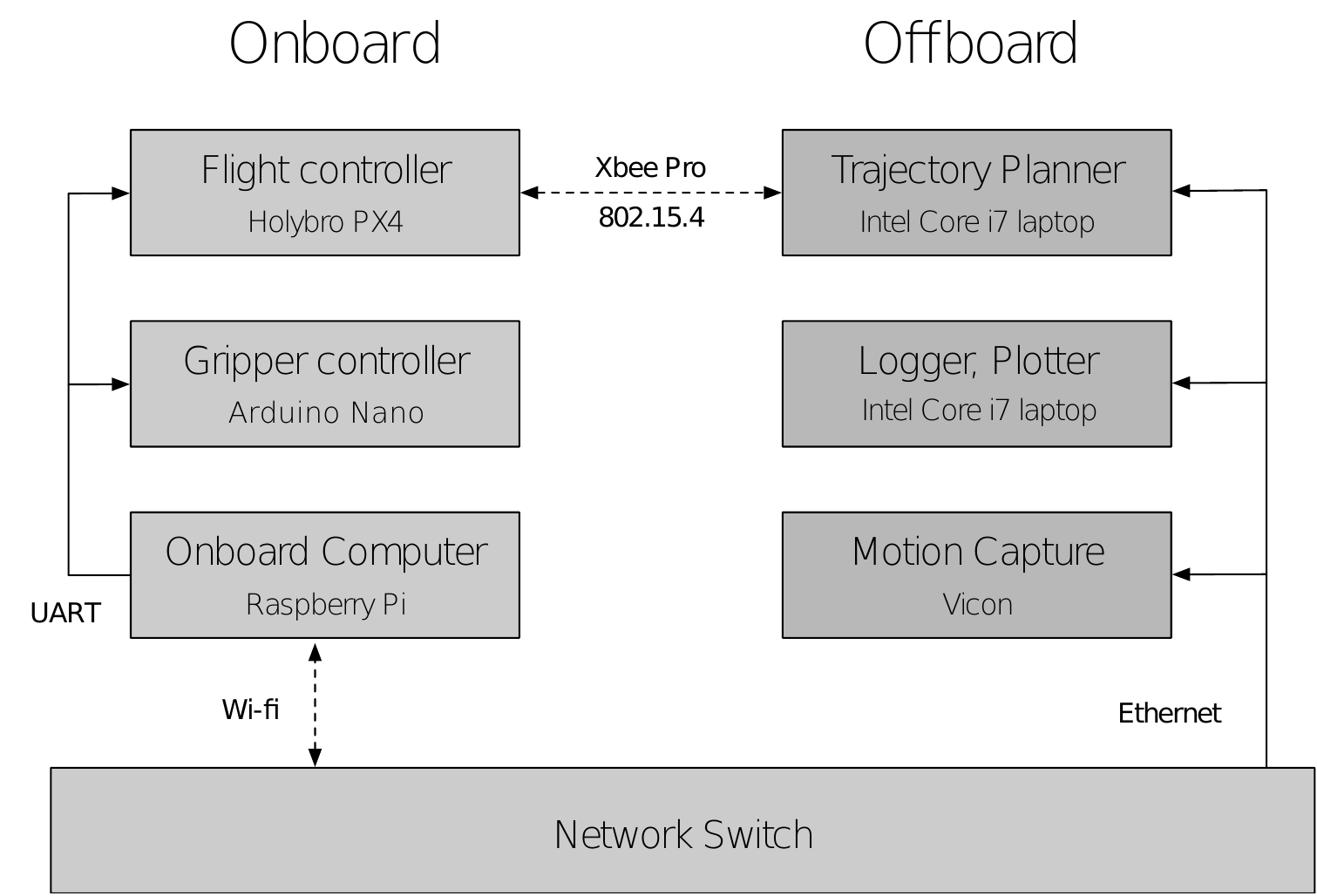}
  \centering
  \\
  \caption{Component architecture of the quadcopter platform. All resource-intensive tasks run offboard to reduce the power draw on the onboard computer and enable convenient development and testing. However, these components can also be moved to the onboard computer for a completely self-contained system with the corresponding performance trade-offs. Onboard communication utilizes universal asynchronous receiver-transmitter (UART) serial connections and the trajectory planner uses the XBee connection. Everything else is connected through the local network.}
\end{figure}

\begin{figure*}[h]
  \centering
  \includegraphics[width=0.9\linewidth]{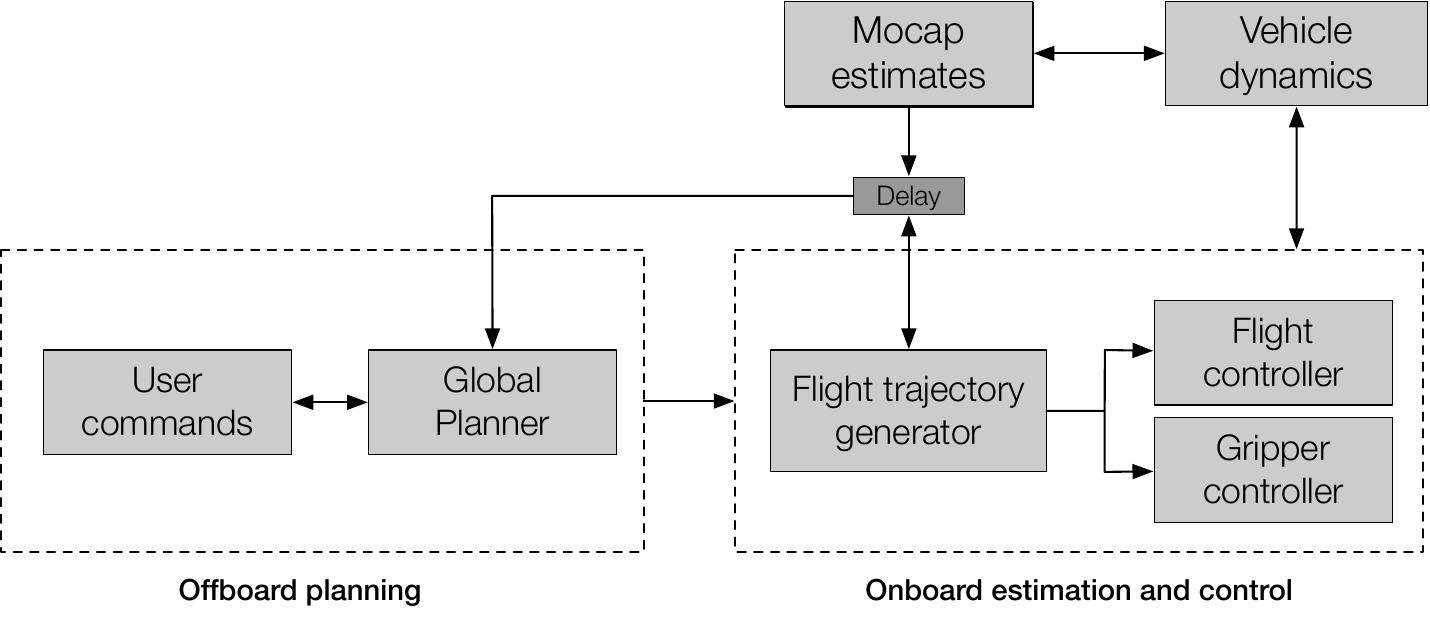}
  \centering
  \\
  \caption{Dataflow architecture of the quadcopter platform. We use a global planner while the minimal-jerk trajectories are computed using the trajectory generator by Mueller \textit{et al.}~\cite{Mueller2015ACE}. The default controller implementation of the PX4 flight controller handles the onboard estimation and control with slightly adjusted PID parameters that are omitted here. All controls are acting based on the pose given by the motion capture system, whereas the delay is accounted for in the PX4's state estimator.}
\end{figure*}

\section{Mechanical Design}\label{mechanical}We explored different state-of-the-art approaches for the design of the gripper: a pneumatic activation \cite{Terrile2021ComparisonOD}, \cite{Amend2012APP}, \cite{Ariyanto2019ThreeFingeredSR} as well as a deformable finger approach (Fin Ray\textsuperscript{\textregistered}) \cite{2011APPLICATIONOF}. We created and evaluated various prototypes based on the following criteria with the corresponding weights: 
\begin{itemize}
    \item Actuation Time (30\%)
    \item Gripping Force (20\%)
    \item Fabrication (20\%)
    \item Total Weight (20\%)
    \item Adaptability/Robustness (10\%)
\end{itemize}

By assigning weights to the different criteria, we were able to compute a score to evaluate the best approach for our use case. Each criterion was given a score from 1 to 10 and then weighted accordingly, yielding a maximum score of 10 for each approach. Table~\ref{table:grippers_comparison} shows such a comparison between the Fin Ray\textsuperscript{\textregistered} and the pneumatic approach. Since the former outperformed the latter in terms of gripping force and ease of fabrication, we proceeded by developing successive iterations of it as prototypes. We then tested our prototypes using only the essential elements on the drone platform. 

\begin{table}
\caption{Score Comparison Between Fin Ray and Pneumatic Prototypes}
\label{table:grippers_comparison}
\begin{center}
\begin{tabular}{ccc}
\toprule
Gripper Type & Fin Ray & Pneumatic\\ 
\midrule
Activation Time & 3 & 3 \\
Gripping Force & 2 & 1.2 \\
Fabrication & 1.6 & 1 \\
Weight & 1.4 & 1.8 \\
Adaptability & 0.8 & 1 \\
\midrule
Total score & 8.8 & 8 \\
\bottomrule
\end{tabular}
\end{center}
\end{table}

Throughout the development of the drone and the gripper, our main focus was on the agility of the quadcopter to allow for actuation and grasping at high velocities. Based on this idea, we built a lightweight version of the components, using the Fused Deposition Modeling (FDM) 3D printing for our prototypes due to its ease of use and speed.

\subsection{Drone Design}
We modified the Holybro X500 platform to achieve our objective of a lightweight design. This was done by adding mounts for the items discussed in Section~\ref{Components}. We arranged the placement of the different components such that the center of gravity remained in the middle of the platform while providing easy access to the various units. 

The Holybro X500 platform consists of carbon fiber plates and tubings as well as plastic connectors. We 3D printed the mounts for the added components and decided to use PLA as the manufacturing material thanks to its ease of printing and mechanical characteristics, which are sufficient for our use. In particular, the temperatures near the PLA parts do not exceed the critical threshold temperature of 55°C, at which PLA begins to soften.

\subsection{Gripper Design}
\subsubsection{Fingers} 
The fingers of the gripper are designed to bend via the Fin Ray\textsuperscript{\textregistered} Effect~\cite{Crooks2016FinRE}---the tip of the structure bends toward the applied force, which enables the fingers to interlock with an object. This additional interlocking mechanism enhances the maximum payload in comparison to a rigid gripper which only works through static friction. Based on previous work on jamming capability~\cite{Elgeneidy2019Characterising3S}, we tilted the ribs, further increasing the generated force.

In addition, we modified the original shape of the Fin Ray\textsuperscript{\textregistered} from an equilateral triangle to a right-angled triangle, which allows for two parallel faces once the gripper is closed. This enlarged contact area enhances the gripping security.

\subsubsection{Actuation and Control}\label{Actuation}
\begin{figure}[h]
\centering
\centerline{\includegraphics[width=\linewidth]{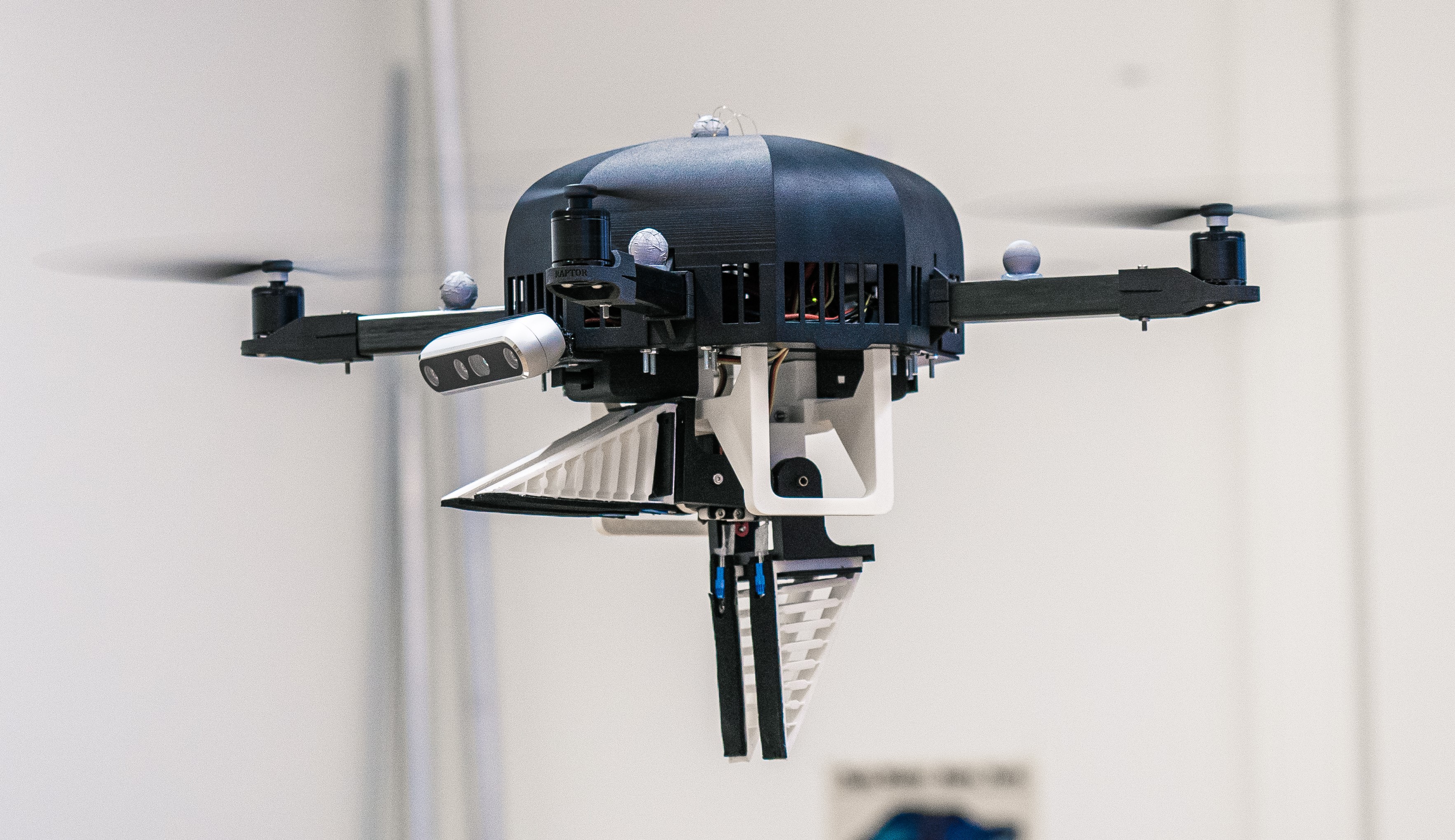}}
\centering
\caption{The last iterations of the gripper fingers. We tested each iteration in a custom-made testbed until we arrived at a gripper iteration with sufficiently compliant fingers to bend around objects and fully utilize the compliant aspect of a soft gripper. The gripper is mounted below the drone.}
\label{gripper}
\end{figure}

For actuation, we use two MG995 servo motors. They are controlled by an Arduino Nano, which receives the angle settings for the servos from the Raspberry Pi 4B. The rotational movement of the gripper is guided on one side by the servo and the other side with a bearing. There is a component that connects the bearing and the rotor in a bow shape to allow for the rotational movement to happen. On each bow, there are two fingers at a distance of \SI{25}{\milli\metre} attached to the bow part. The fingers are in parallel so that they align with their counterpart. There are two servos and therefore two bows and four fingers.

\subsubsection{Mount}
We mounted the gripper construction on the drone via two parallel tubes that are fixed due to friction with a rubber piece. The counterpart of the gripper has two holes to accommodate the tubes. Further, there are two rectangular holes in the mount for the servos. The servos can be fixed to the mount using four screws. Additionally, one part of the bearing of the bow from Section~\ref{Actuation} is also a part of the mount. Figure~\ref{gripper} illustrates the complete gripper assembly as well as the actuation configuration discussed in the previous section.

\subsubsection{Fabrication}
We used FDM 3D printing for fabrication due to its speed in manufacturing new prototypes. We chose PLA for the rigid parts of the gripper because of its ease of use. For the fingers, we decided to use Extrudr Flex Semisoft filament material due to its similarity to Ninjaflex, which is commonly used for 3D printed fingers \cite{Ali2019BiologicallyIG}.

\section{Trajectory Generation and Control}\label{control}
\subsubsection{Onboard Control}
A Raspberry Pi 4B and a Pixhawk 4 (PX4) flight controller form the drone's primary components. The Raspberry Pi is mainly responsible for onboard computing and handles the communication between the onboard components and the local network. It accesses data from the motion capture system via Fast DDS, which is then forwarded to the PX4 flight controller over a serial connection using UART. The state estimator in the PX4 allows for the compensation of small latencies by a user-configurable parameter. We use the low-level attitude and attitude rate controllers that are available by default on the PX4. The attitude controller is a nonlinear PD controller and runs at \SI{500}{\hertz}. The attitude rate controller is a tightly tuned PID controller running at \SI{1000}{\hertz}.

\subsubsection{Offboard Control}
Our system architecture allows us to run computationally heavy tasks in real time on offboard computers with more computational power. The motion capture data published on the local network can be accessed by the offboard computers through Fast DDS. Using the pose of the drone and a target object, waypoints are computed in real time at a frequency of \SI{50}{\hertz}. These high-level commands are then sent directly to the PX4 over an additional XBee wireless radio link. Additional applications for logging, plotting, and simple high-level commands are also run on offboard computers to allow for the rapid testing of new software and hardware components. These applications are able to communicate via Fast DDS using either shared memory if run on the same computer or the User Datagram Protocol (UDP) if run on different computers.

\subsubsection{Gripper Control}
Attached to the bottom of the drone is a gripper module consisting of two arms actuated by a high-speed servomotor each, which are in turn controlled by an Arduino Nano microcontroller. Any high-level commands sent by the offboard controls through the local network via Fast DDS are first received by the Raspberry Pi before being forwarded to the microcontroller using the UART serial communication protocol.

\section{Experiments}\label{experiments}
\subsection{Setup}
\begin{figure}[h]
\centering
\centerline{\includegraphics[width=\linewidth]{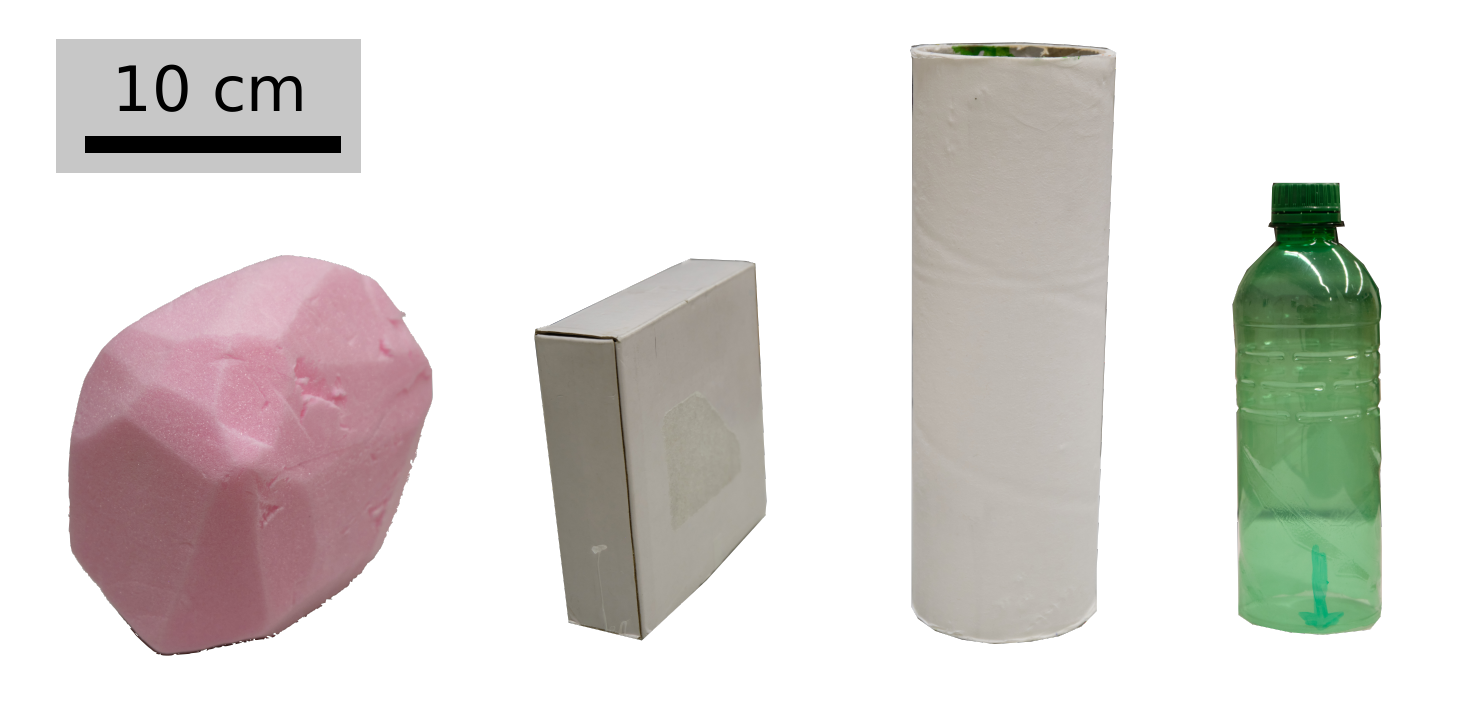}}
\centering
\caption{Dimensions of the testing objects relative to each other. Our motivation for the choices was to expose our gripper to a variety of different geometries and the associated challenges for aerial grasping.\label{testing_objects}}
\label{fig7}
\end{figure}

We tested RAPTOR with four different objects: 1) an irregularly shaped styrofoam object, 2) a small planar cardboard box, 3) a paper roll, and 4) a \SI{0.5}{\liter} PET bottle. Figure~\ref{testing_objects} shows the different object geometries used in our experiments with the drone and the gripper and Table~\ref{object_properties} highlights the different dimensions and masses associated with each object. We placed these objects on a custom-made stand made out of aluminum profiles. The stand ensured that object positioning remained consistent both throughout and during the attempts. For each object, we performed a total of 36 grasps. We performed all grasping attempts over two consecutive days with the same gripper as seen in Figure~\ref{gripper}. Between these attempts, we only landed the drone to exchange the battery to ensure that the condition of the drone remained the same.

\begin{table}
\caption{Object Properties}
\label{object_properties}
\begin{center}
\begin{tabular}{ccc}
\toprule
Object & Mass [g] &  Dimensions \( l \times w \times h \) [\si{\centi\metre}]\\
\midrule
Styrofoam object &  27 & \( 9 \times 14 \times 13 \) \\
Cardboard box &  34 & \( 3 \times 12 \times 12 \) \\
Paper roll &  86 & \( 8 \times 8 \times 24 \) \\
PET Bottle &  17 & \( 6 \times 6 \times 19 \) \\
\bottomrule
\end{tabular}
\end{center}
\end{table}

\begin{table}
\caption{Success Rate and Average Grasping Velocity for the Different Geometries}
\label{success_rates}
\begin{center}
\begin{tabular}{ccc}
\toprule
Object & Success rate & Average grasping velocity [\si{\metre\per\second}]\\ 
\midrule
Styrofoam object &  100\% & $1.05 \pm 0.04 $\\
Cardboard box &  94\% &$0.99 \pm 0.06$\\
Paper roll &  75\% &$1.01 \pm 0.05$\\
PET Bottle &  61\% &$1.00 \pm 0.06$\\
\bottomrule
\end{tabular}
\end{center}
\end{table}

\begin{figure}[h]
\centering
\centerline{\includegraphics[width=\linewidth]{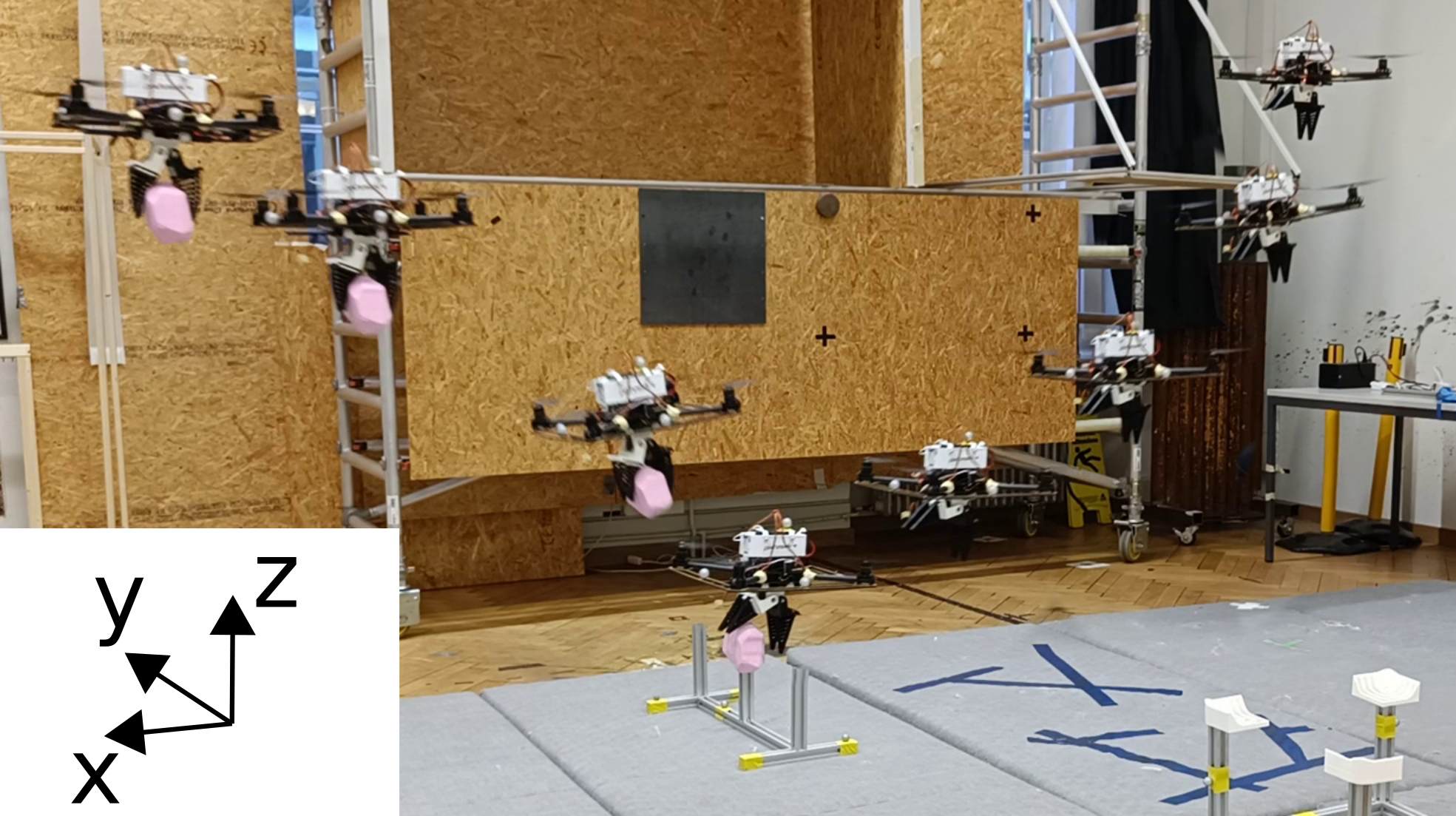}}
\centering
\caption{Experimental setup including the coordinate system of the motion capture. In this setup, the x-axis corresponds to the longitudinal translation to the object while the y-axis describes the lateral translation relative to the object.\label{fig:grasp_setup}}
\end{figure}

\subsection{Results}
Table~\ref{success_rates} shows the success rates and average velocity as well as the standard deviation for successful grasping throughout the swooping motion to pick up the object for the 36 swooping motions per object. We define each swooping motion to begin at a distance of 2 meters away from the object and end 2 meters behind the object. Based on these assumptions, we computed the average speed with the time needed to complete the full motion and the distance traveled. We can observe that for all the object geometries tested, the average grasping velocity is approximately \SI{1}{\metre\per\second}.

\begin{figure}[h]
\centering
\centerline{\includegraphics[width=\linewidth]{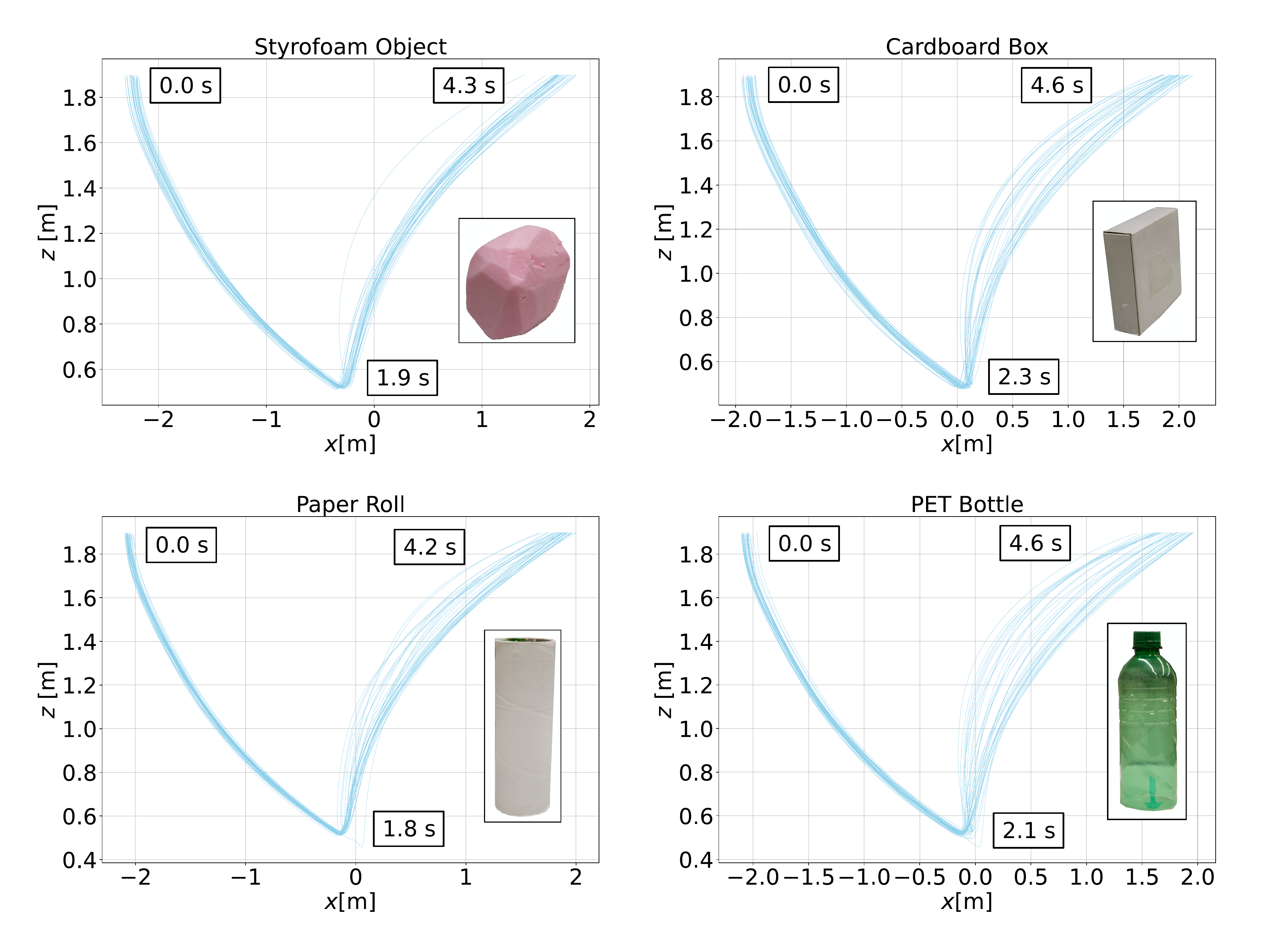}}
\centering
\caption{Flight logs of the quadcopter swooping trajectory in the xz-plane for different objects with 36 grasps attempted for each object. We observe a strong consistency for the styrofoam and the paper roll with only a few outliers in the trajectories. The data for the box and the PET bottle show a tendency to have more vertical trajectories. This can be explained by the box and the PET bottle having a longer travel distance until they are fully lifted from the stand, which causes the controller to produce this overshoot. \label{swoop_logs}}
\end{figure}

\begin{figure}[h]
\centering
\centerline{\includegraphics[width=\linewidth]{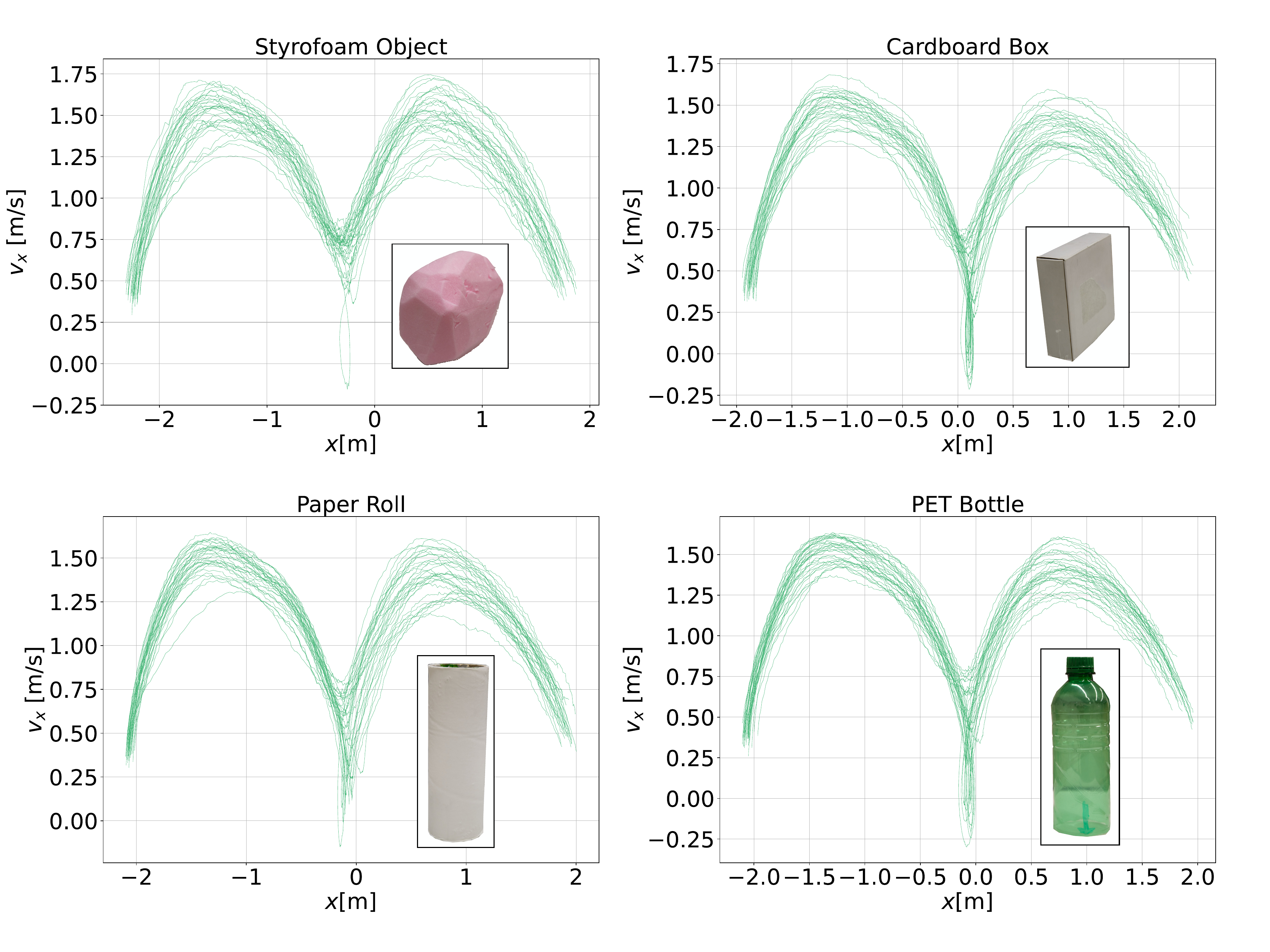}}
\centering
\caption{The pickup speed depends on which object the drone is picking up. For instance, for the styrofoam object, we can see a consistent minimum speed at the grasp of just below \SI{0.5}{\metre\per\second}. For the other objects, this minimum speed decreases and the drone momentarily almost comes to a full stand. \label{swoop_velocity}}
\end{figure}

Figure~\ref{swoop_logs} and Figure~\ref{swoop_velocity} illustrate the trajectories and velocities of the drone during pick-up attempts respectively. Moreover, they show the amount of vertical travel required to lift an object from the stand. The styrofoam object requires the least vertical travel to be lifted from the stand while the other objects require a larger vertical travel of approximately \SI{4}{\centi\metre}. As we are not modelling this effect in our control scheme, if the drone gets stuck in the swooping motion trying to lift up an object, this can both reduce the momentary velocity of the drone and cause an overshoot following a successful grasping of the object from the stand.

In addition, we notice a very reliable grasping performance for the styrofoam object and the box, with the drone consistently able to pick up the object while moving at an average speed that is five times the speed reported in previous works for an actively controlled gripper \cite{Fishman2021DynamicGW}. Performance degrades as we transition to smaller objects such as the paper roll and the PET bottle due to reference tracking noise in the xy-plane. There is a clear correlation between the width of the object and the grasping success of RAPTOR, which can be seen with rigid grippers as well~\cite{Stewart2022HowTS}. Notably, the other dimensions of the object do not seem to have an impact on success. This demonstrates the advantages of compliant soft grippers. They are largely invariant to the surface smoothness of the object that is to be grasped---the styrofoam object has the roughest surface while the box has a very smooth surface. Even when the quadcopter only achieves a partial grasp with two fingers, it maintains a steady grip with no noticeable degradation in trajectory tracking. Additionally, based on our results, we can expand previous works to smaller objects and achieve higher grasping efficacy for our two largest objects~\cite{Fishman2021DynamicGW}.

\subsection{Limitations}
\subsubsection{Failure Modes}
We also observe that the failures are primarily due to the controller of the quadcopter not tracking the reference accurately enough in the xy-plane and missing the object by flying past it without making contact. Whenever it achieved a sufficient reference tracking such that at least two fingers made contact with the object, the quadcopter could complete a successful grasp. Objects did not slip out of the gripper once they had been grasped, even if the grasp was only with one pair of fingers. This behavior could be observed across all objects with their respective surfaces and shapes. Grasp mistiming is also not an issue, since the gripper is actuated only if the drone is in the close vicinity of the object that is being grasped.

\subsubsection{Reference Tracking}
\begin{figure}[h]
\centering
\centerline{\includegraphics[width=0.99\linewidth]{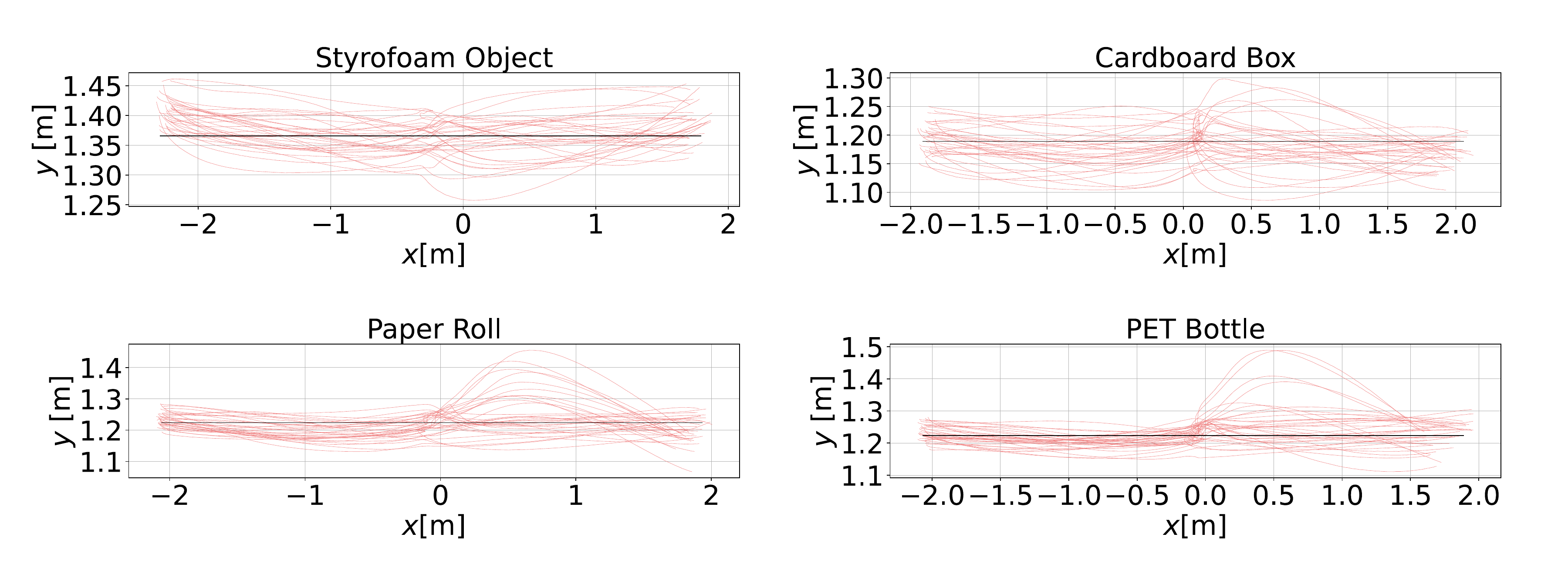}}
\centering
\caption{The y-coordinate (lateral translation with respect to the object) of the drone during the pick-up attempts of the different objects. The trajectory varies significantly during different attempts. The approach of the object is consistently distributed across approximately \SI{10}{\centi\metre}. \label{swoop_y_deviation}}
\end{figure}

Figure~\ref{swoop_y_deviation} shows the plots for the planar coordinates of the drone in more detail. Here, the deviation for the grasping attempts for each object is clearly visible. Given that the width of the gripper is approximately \SI{6}{\centi\metre} and in most cases, we only need to have contact with one of the two pairs of fingers for a successful grasp, this allows for a deviation that should be approximately half of the width of the object. Considering Table~\ref{object_properties} and the width of the objects, a deviation of only a few centimeters towards one side of the objects will already have a major effect on grasping efficacy, particularly for the objects with the smallest width.

\subsubsection{Ground Effect}
Moreover, near the ground, aerodynamic effects such as the ground effect \cite{Kan2019AnalysisOG} add additional factors of uncertainty to the drone dynamics which the controller does not account for. Previous works have attempted to model the ground effect \cite{Bernard2018GroundEA} and even account for it with a controller \cite{Fishman2021DynamicGW}, but there have been no controllers proposed to counter the effect for the dynamic grasping of objects. Recent learning-based approaches \cite{Shi2019NeuralLS} have been shown to reduce tracking uncertainty, but they are currently unable to generalize to different environments. 

\begin{figure*}[h]
\vspace{3mm}
  \centering
  \includegraphics[width=0.99\linewidth]{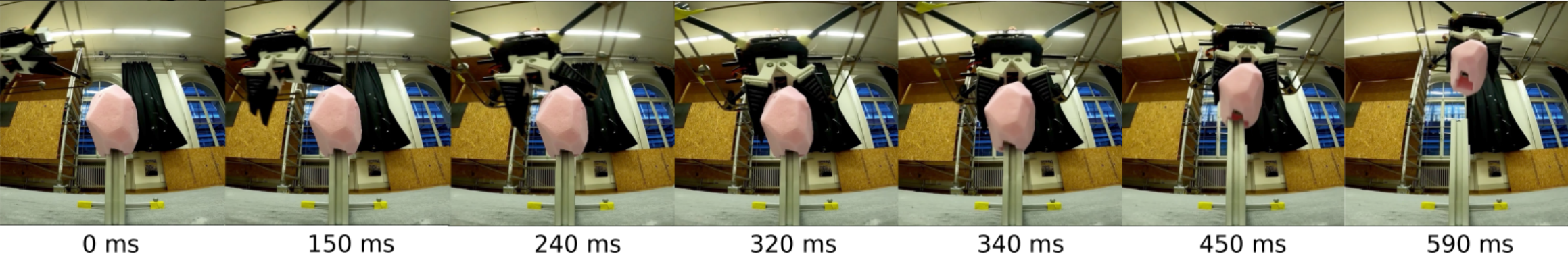}
  \centering
  \caption{Grasping sequence for our quadcopter platform. The drone approaches with the gripper open and as soon as the drone has reached the target position, the gripper closes, lifting up the object.}
\end{figure*}

\section{Conclusion}\label{conclusion}
With RAPTOR, we demonstrate a platform for robust dynamic aerial grasping. Using soft grippers, we can successfully grasp objects of different geometries and surface finishes, even at high speeds where traditional rigid grippers would fail. We describe our system architecture and mechanical design in detail, drawing comparisons between our custom system components such as our custom Fast DDS wrapper and existing solutions for building similar robotic systems as well as motivating our design decisions, particularly for the soft gripper. We then evaluate our system using real-world experiments that validate our design decisions and show significant improvements compared to previous works. In particular, we were able to show robust grasping at high speeds---with average velocities of \SI{1}{\metre\per\second}---while our quadcopter could also grasp smaller objects that have additional requirements for reference tracking, showing potential for a wider range of applications.

While there are many frontiers in dynamic aerial grasping that require further research, we see the more precise modeling of aerodynamic ground effects for reference tracking as the main challenge to consistently picking up small objects, especially when the drone is operating close to surfaces. Moreover, future work will require to address the challenges associated with making the platform independent of motion capture systems while exploring the picking up of objects without a stand due to the additional force caused by the propeller thrust. An accurate and computationally feasible model of aerodynamic effects will therefore be a major milestone for a more robust dynamic aerial grasping.

%%%%%%%%%%%%%%%%%%%%%%
% \newpage
\bibliography{root.bib}{}

@article{Fishman2021DynamicGW,
  title={Dynamic Grasping with a "Soft" Drone: From Theory to Practice},
  author={Joshua Fishman and Samuel Ubellacker and Nathan Hughes and Luca Carlone},
  journal={2021 IEEE/RSJ International Conference on Intelligent Robots and Systems (IROS)},
  year={2021},
  pages={4214-4221}
}

@article{Fishman2021ControlAT,
  title={Control and Trajectory Optimization for Soft Aerial Manipulation},
  author={Joshua Fishman and Luca Carlone},
  journal={2021 IEEE Aerospace Conference (50100)},
  year={2021},
  pages={1-17}
}

@article{Mueller2015ACE,
  title={A Computationally Efficient Motion Primitive for Quadrocopter Trajectory Generation},
  author={Mark Wilfried Mueller and Markus Hehn and Raffaello D’Andrea},
  journal={IEEE Transactions on Robotics},
  year={2015},
  volume={31},
  pages={1294-1310}
}

@inproceedings{Thomas2013AvianInspiredGF,
  title={Avian-Inspired Grasping for Quadrotor Micro UAVs},
  author={Justin R. Thomas and Joe Polin and Koushil Sreenath and Vijay R. Kumar},
  year={2013}
}

@inproceedings{Quigley2009ROSAO,
  title={ROS: an open-source Robot Operating System},
  author={Morgan Quigley},
  booktitle={ICRA 2009},
  year={2009}
}

@article{Crooks2016FinRE,
  title={Fin Ray{\textregistered} Effect Inspired Soft Robotic Gripper: From the RoboSoft Grand Challenge toward Optimization},
  author={Whitney Crooks and Gabrielle Davis Vukasin and Maeve O'Sullivan and William C. Messner and Chris Rogers},
  journal={Frontiers Robotics AI},
  year={2016},
  volume={3},
  pages={70}
}

@article{Elgeneidy2019Characterising3S,
  title={Characterising 3D-printed Soft Fin Ray Robotic Fingers with Layer Jamming Capability for Delicate Grasping},
  author={Khaled Elgeneidy and Peter Lightbody and Simon Pearson and Gerhard Neumann},
  journal={2019 2nd IEEE International Conference on Soft Robotics (RoboSoft)},
  year={2019},
  pages={143-148}
}

@article{Amend2012APP,
  title={A Positive Pressure Universal Gripper Based on the Jamming of Granular Material},
  author={John R. Amend and Eric Brown and Nicholas Rodenberg and Heinrich M. Jaeger and Hod Lipson},
  journal={IEEE Transactions on Robotics},
  year={2012},
  volume={28},
  pages={341-350}
}

@article{Terrile2021ComparisonOD,
  title={Comparison of Different Technologies for Soft Robotics Grippers},
  author={Silvia Terrile and Miguel Arg{\"u}elles and Antonio Barrientos},
  journal={Sensors (Basel, Switzerland)},
  year={2021},
  volume={21}
}

@article{Ariyanto2019ThreeFingeredSR,
  title={Three-Fingered Soft Robotic Gripper Based on Pneumatic Network Actuator},
  author={Mochammad Ariyanto and Munadi Munadi and Joga Dharma Setiawan and Dedi Mulyanto and Tanto Nugroho},
  journal={2019 6th International Conference on Information Technology, Computer and Electrical Engineering (ICITACEE)},
  year={2019},
  pages={1-5}
}

@article{MantiMariangela2015ABS,
  title={A Bioinspired Soft Robotic Gripper for Adaptable and Effective Grasping},
  author={Mariangela Manti and Taimoor Hassan and Giovanni Passetti and Nicol{\`o} D'Elia and Cecilia Laschi and Matteo Cianchetti},
  journal={Soft robotics},
  year={2015},
  volume={2},
  pages={107-116}
}

@article{Kronauer2021LatencyOO,
  title={Latency Overhead of ROS2 for Modular Time-Critical Systems},
  author={Tobias Kronauer and Joshwa Pohlmann and Maximilian Matth{\'e} and Till Smejkal and Gerhard P. Fettweis},
  journal={ArXiv},
  year={2021},
  volume={abs/2101.02074}
}

@article{Rus2015DesignFA,
  title={Design, fabrication and control of soft robots},
  author={Daniela Rus and Michael Thomas Tolley},
  journal={Nature},
  year={2015},
  volume={521},
  pages={467-475}
}

@article{GeorgeThuruthel2018ControlSF,
  title={Control Strategies for Soft Robotic Manipulators: A Survey.},
  author={Thomas George Thuruthel and Yasmin Ansari and Egidio Falotico and Cecilia Laschi},
  journal={Soft robotics},
  year={2018},
  volume={5 2},
  pages={
          149-163
        }
}

@article{Shi2019NeuralLS,
  title={Neural Lander: Stable Drone Landing Control Using Learned Dynamics},
  author={Guanya Shi and Xichen Shi and Michael O'Connell and Rose Yu and Kamyar Azizzadenesheli and Anima Anandkumar and Yisong Yue and Soon-Jo Chung},
  journal={2019 International Conference on Robotics and Automation (ICRA)},
  year={2019},
  pages={9784-9790}
}

@article{Ali2019BiologicallyIG,
  title={Biologically Inspired Gripper Based on the Fin Ray Effect},
  author={Md. Hazrat Ali and Asset Zhanabayev and Samat Khamzhin and Kainar Mussin},
  journal={2019 5th International Conference on Control, Automation and Robotics (ICCAR)},
  year={2019},
  pages={865-869}
}

@inproceedings{Bernard2018GroundEA,
  title={Ground Effect Analysis for a Quadrotor Platform},
  author={Davide Del Cont Bernard and Mattia Giurato and Fabio Riccardi and Marco Lovera},
  year={2018}
}

@article{Bodie2019AnOA,
  title={An Omnidirectional Aerial Manipulation Platform for Contact-Based Inspection},
  author={Karen Bodie and Maximilian Brunner and Michael Pantic and Stefan Walser and Patrick Pf{\"a}ndler and Ueli M. Angst and Roland Y. Siegwart and Juan I. Nieto},
  journal={ArXiv},
  year={2019},
  volume={abs/1905.03502}
}

@article{Falanga2019TheFD,
  title={The Foldable Drone: A Morphing Quadrotor That Can Squeeze and Fly},
  author={Davide Falanga and Kevin Kleber and Stefano Mintchev and Dario Floreano and Davide Scaramuzza},
  journal={IEEE Robotics and Automation Letters},
  year={2019},
  volume={4},
  pages={209-216}
}

@inproceedings{Garimella2018NonlinearMP,
  title={Nonlinear Model Predictive Control of an Aerial Manipulator using a Recurrent Neural Network Model},
  author={Gowtham Garimella and Mathew Sheckells},
  year={2018}
}

@article{Haque2014AutonomousQF,
  title={Autonomous Quadcopter for product home delivery},
  author={Md R. Haque and M. Muhammad and D. Swarnaker and Md Arifuzzaman},
  journal={2014 International Conference on Electrical Engineering and Information \& Communication Technology},
  year={2014},
  pages={1-5}
}

@article{Kamel2016DesignAM,
  title={Design and modeling of dexterous aerial manipulator},
  author={Mina Kamel and Kostas Alexis and Roland Y. Siegwart},
  journal={2016 IEEE/RSJ International Conference on Intelligent Robots and Systems (IROS)},
  year={2016},
  pages={4870-4876}
}

@article{King2018DesignFA,
  title={Design. Fabrication, and Evaluation of Tendon-Driven Multi-Fingered Foam Hands},
  author={Jonathan P. King and Dominik Bauer and Cornelia Schlagenhauf and Kai-Hung Chang and Daniele Moro and Nancy S. Pollard and Stelian Coros},
  journal={2018 IEEE-RAS 18th International Conference on Humanoid Robots (Humanoids)},
  year={2018},
  pages={1-9}
}

@article{Hassan2015DesignAD,
  title={Design and development of a bio-inspired, under-actuated soft gripper},
  author={Taimoor Hassan and Mariangela Manti and Giovanni Passetti and Nicol{\`o} d'Elia and Matteo Cianchetti and Cecilia Laschi},
  journal={2015 37th Annual International Conference of the IEEE Engineering in Medicine and Biology Society (EMBC)},
  year={2015},
  pages={3619-3622}
}

@article{Crooks2017PassiveGI,
  title={Passive gripper inspired by Manduca sexta and the Fin Ray{\textregistered} Effect},
  author={Whitney Crooks and Shane Rozen-Levy and Barry Andrew Trimmer and Chris Rogers and William C. Messner},
  journal={International Journal of Advanced Robotic Systems},
  year={2017},
  volume={14}
}

@inproceedings{2011APPLICATIONOF,
  title={APPLICATION OF FINRAY EFFECT APPROACH FOR PRODUCTION PROCESS AUTOMATION},
  author={},
  year={2011}
}

@article{Kan2019AnalysisOG,
  title={Analysis of Ground Effect for Small-Scale UAVs in Forward Flight},
  author={Xinyue Kan and Justin R. Thomas and Hanzhe Teng and Herbert G. Tanner and Vijay R. Kumar and Konstantinos Karydis},
  journal={IEEE Robotics and Automation Letters},
  year={2019},
  volume={4},
  pages={3860-3867}
}

@article{McLaren2019APC,
  title={A Passive Closing, Tendon Driven, Adaptive Robot Hand for Ultra-Fast, Aerial Grasping and Perching},
  author={Andrew I. McLaren and Zak Fitzgerald and Geng Gao and Minas Liarokapis},
  journal={2019 IEEE/RSJ International Conference on Intelligent Robots and Systems (IROS)},
  year={2019},
  pages={5602-5607}
}

@article{Tanaka2019HighspeedUD,
  title={High-speed UAV Delivery System with Non-stop Parcel Handover Using High-speed Visual Control},
  author={Satoshi Tanaka and Taku Senoo and Masatoshi Ishikawa},
  journal={2019 IEEE Intelligent Transportation Systems Conference (ITSC)},
  year={2019},
  pages={4449-4455}
}

@article{Stewart2022HowTS,
  title={How to Swoop and Grasp Like a Bird With a Passive Claw for a High-Speed Grasping},
  author={William Stewart and Enrico Ajanic and Matthias Muller and Dario Floreano},
  journal={IEEE/ASME Transactions on Mechatronics},
  year={2022}
}

@article{Shan2020ModelingAA,
  title={Modeling and analysis of soft robotic fingers using the fin ray effect},
  author={Xiaowei Shan and Lionel Birglen},
  journal={The International Journal of Robotics Research},
  year={2020},
  volume={39},
  pages={1686 - 1705}
}
\bibliographystyle{IEEEtran} 

\end{document}